# A Realistic Model Reference Computed Torque Control Strategy for Human Lower Limb Exoskeletons

**SK Hasan**
Department of Mechanical and Manufacturing Engineering
Miami University
Oxford, OH 45056

**Corresponding Author:**
  SK Hasan
  hasanSk@miamioh.edu
  Mechanical and Manufacturing Engineering Dept.
  Miami University
  Oxford, OH 45056
  Phone: 513-529-0805

# Contents

# Abstract

Exoskeleton robots have become a promising tool in neurorehabilitation, offering effective physical therapy and recovery monitoring. The success of these therapies relies on precise motion control systems. Although computed torque control based on inverse dynamics provides a robust theoretical foundation, its practical application in rehabilitation is limited by its sensitivity to model accuracy, making it less effective when dealing with unpredictable payloads. To overcome these limitations, this study introduces a novel model reference computed torque controller that accounts for parametric uncertainties while optimizing computational efficiency. A dynamic model of a seven-degree-of-freedom human lower limb exoskeleton is developed, incorporating a realistic joint friction model to accurately reflect the physical behavior of the robot. To reduce computational demands, the control system is split into two loops: a slower loop that predicts joint torque requirements based on input trajectories and robot dynamics, and a faster PID loop that corrects trajectory tracking errors. Coriolis and centrifugal forces are excluded from the model due to their minimal impact on system dynamics relative to their computational cost. Experimental results show high accuracy in trajectory tracking, and statistical analyses confirm the controller's robustness and effectiveness in handling parametric uncertainties. This approach presents a promising advancement for improving the stability and performance of exoskeleton-based neurorehabilitation.

# Keywords

Realistic Model Reference Computed Torque Control, Exoskeleton Robot, Lower Extremity Dynamic Model

# 1. Introduction

Physical disability significantly impairs a person's mobility and physical capabilities, affecting both individuals and society. Nearly a billion people worldwide face challenges in accessing adequate treatment due to limited resources. Exoskeleton robots offer a promising solution for rehabilitation, providing personalized physiotherapy and continuous performance assessment throughout recovery. This technology holds great potential for enhancing physical therapy and improving neurorehabilitation outcomes for individuals with disabilities affecting both upper and lower extremities.

Effective rehabilitation with exoskeleton robots requires precise maneuverability, which depends on an integrated control system combining sensing, actuation, and computation units that communicate via standard protocols and are driven by control algorithms. The high-performance control execution server functions as the brain of the rehabilitation robot, enabling it to provide passive, active, active-assist, and active-resist therapies. Sensors monitor the robot and facilitate interaction with the user, while actuators, functioning like muscles, move the mechanical components. However, even advanced sensors and actuators are insufficient without efficient control algorithms to ensure proper robot maneuverability. Researchers have developed various linear and nonlinear control algorithms to enhance maneuverability and optimize the delivery of diverse physical therapies.

Table 1 summarizes the features of recent human lower limb exoskeleton robots, detailing their actuated joints, control algorithms, and key characteristics.

Table 1: Overview of existing rehabilitation robots, including their degrees of freedom, sensing, actuation, and control systems.

| # | Device name | Actuated joints | Control algorithm | Remarks | Ref. |
|---|---|---|---|---|---|
| 1. | EXPO and SUBAR | Hip: F-E (A) Knee: F-E (A) Ankle: (U) Double legs | EXPO utilizes a Fuzzy logic controller, while SUBAR is developed with an Impedance controller | The use of the impedance controller in SUBAR offers greater user comfort compared to EXPO | [1], [2] |
| 2. | Lokomoat | Pelvis: V.M. (U) Hip: F-E (A) | Patient-Driven Motion Reinforcement (PDMR) | The PDMR method enables patients to walk at their | [3], [4] |

| | | Knee: F-E (A)<br>Ankle: F-E (U)<br>Double legs | techniques were used to implement the Hybrid Force Position Control System | preferred speed and gait pattern | |
|---|---|---|---|---|---|
| 3. | Lopes | Pelvis: L-R, F-B<br>Hip: F-E(A)<br>A-A (A)<br>Knee: F-E (A)<br>Double legs | An impedance control scheme was implemented | The impedance controller enabled the creation of a virtual therapist action | [5], [6] |
| 4. | ALEX | Trunk: 3 DOF<br>Hip: F-E (A), A-A(U)<br>Knee: F-E (A)<br>Ankle: F-E(U)<br>Single leg | A force field controller was implemented, allowing the system to apply appropriate force to follow the desired trajectory | The force field controller enabled the development of the assist-as-needed control technique, which is crucial for active and active-assist types of physical therapy | [7], [8], [9], [10] |
| 5. | HAL | Hip: F-E (A)<br>Knee: F-E (A)<br>Ankle: F-E (U)<br>Single leg | HAL single-leg version operates using a hybrid controller that combines Cybernetic Voluntary Control (CVC) and Cybernetic Autonomous Control (CAC). CVC offers physical support based on the user's voluntary muscle activity, while CAC functions using pre-recorded trajectories | CVC is effective when the user has strong voluntary muscle signals, while CAC is more suitable when those signals are weak. This allows the system to accommodate all types of patients | [11], [12], [13] |
| 6. | REWALK | Hip: F-E(A)<br>Knee: F-E(A)<br>Foot: F-E(U)<br>Double legs | It operates using pre-recorded trajectories, with a tilt sensor determining the trunk angle to select the most appropriate trajectory for the user's condition | The literature lacks a detailed description of the low-level controller | [14], [15] |
| 7. | ELEGS | Hip: A-A (U)<br>F-E (A)<br>Knee: F-E (A)<br>Ankle: F-E (U) | A finite state machine is used to maneuver a series of states. | Different controllers were developed for different states. | [16], [17] |
| 8. | Vanderbilt Exoskeleton | Hip: F-E (A)<br>Knee: F-E (A)<br>Double legs | Runs based on preprogrammed trajectories. It has settings for different modes like sitting to stand, walk, stair ascent/descent | The literature lacks a detailed description of the low-level controller | [18] |
| 9. | ATLAS | Hip: F-E (A)<br>Knee: F-E (A)<br>Ankle: F-E(U) | It was developed by integrating a finite state machine with a PD controller, with specific proportional and derivative gains, along with defined entry and exit conditions for each state | The finite state machine, combined with varying gains, functions similarly to a gain scheduling mechanism | [19] |
| 10. | MINA | Hip: F-E (A) Knee: F-E (A) Ankle: F-E(U) Double legs | MINA operates using a PD controller and functions in two phases: the recording phase, where trajectories are collected from healthy subjects, and the running phase, during which the robot follows the pre-recorded trajectories | The use of a PD controller for a type I system is well justified, as it ensures both stability and accuracy | [20] |
| 11. | Mind walker | Hip: A-A (A) F-E(A) Knee: F-E(A) Ankle: F-E (U) Double legs | A joint impedance controller is integrated with a Finite State Machine, where state transitions are triggered by shifts in the center of mass. | This approach is highly effective for controlling walking assistance in lower limb exoskeleton robots. The inclusion of an impedance controller allows the robot to easily match its impedance with the user. | [21] |
| 12. | Walking assistance lower limb exoskeleton | Hip: F-E (A) A-A (U) Knee: F-E (A) Ankle: U Double Legs | The walking assistance robot operates using a finite state machine, with state transitions triggered by the location of the Center of Pressure, determined through data from the inclinometer | The inclinometer mounted on the backbone measures the torso angle. | [22] |
| 13. | IHMC mobility assist exoskeleton | Hip: F-E (A), A-A (A), R-R (U) Knee: F-E (A) Ankle: F-E (U) Double Legs | System operates based on torque and position control, with a PD controller used in both cases. | The PD controller is employed for both position and torque control, with greater emphasis placed on system robustness than on tracking accuracy | [23] |
| 14. | Lower-limb power assist exoskeleton | Hip: F-E(A) Knee: F-E(A) Ankle: U | A PI velocity control loop is placed within a PI torque control loop. | More emphasis was placed on accuracy rather than the robustness of the control system. | [24] |
| 15. | ABLE | Hip: F-E Knee: F-E Ankle: F-E | Powered by a PD controller, the mobile platform, lower limb orthosis, and telescopic crutch operate in synchrony | More focus was given on the stability and disturbance rejection rather than tracking accuracy. | [25] |
| 16. | Nurse robot suit | Supports shoulder, waist, legs | PID control technique | PID control algorithms were employed to achieve a balance between stability and accuracy | [26], [27] |
| 17. | BLEEX | Hip: F-E (A), A-A (A), R (U) Knee: F-E (A) Ankle: A-A(U), F-E(A) R | The BLEEX robot operates with a hybrid controller, consisting of two separate controllers: one for the swing phase and another for the stance phase. The | The combination of the position controller and the positive feedback-based sensitivity controller performed efficiently, | [28], [29] |

| | | | swing phase demands high velocity with low torque, while the stance phase requires low velocity with high torque | resulting in minimal tracking error | |
|---|---|---|---|---|---|
| 18. | CUHK-Exo | Hip: F-E (A), R Knee: F-E (A) Ankle: F-E(P) | At the lower level, a PD controller is utilized, while the upper level employs the Offline Design and Online Modification (ODOM) control technique | Accurately calculating the center of pressure is impractical, but the use of the ODOM adaptation method enhances its functionality | [30], [31] |
| 19. | Xor | Hip: F-E (A), R(U) Knee: F-E (A) Ankle: F-E(P), A-A(U) | The system operates on a hybrid driving concept, with pneumatic artificial muscles serving as gravity balancers and an electric motor acting as a compensator | A comparison with the PD controller demonstrates the effectiveness of the proposed controller. | [32] |

F-E→ Flexion-Extension, A-A →Abduction-Adduction, L-R→ Left-Right, F-B→ Forward-Backward, A→ Actuated, U→ Unactuated, VM→ Vertical Movement

Table 1 illustrates that both linear and nonlinear control algorithms are utilized to address nonlinear robot dynamics. Most nonlinear control algorithms rely on a mathematical model of the robot's dynamics, making it essential to create an accurate model as part of the control architecture. Although PD and PID techniques do not strictly require a model, having one aids in gain selection and performance evaluation. Developing an accurate robot model is challenging, but feedback control mechanisms offer some protection against modeling inaccuracies. The acceptable modeling error range varies by control technique, with some algorithms being sensitive to errors while others are more robust against uncertainties [33].

This paper presents a 7-DOF dynamic model of the human lower extremity developed using the Lagrange energy method. To control this model, a novel Realistic Model Reference Computed Torque Controller (RMRCTC) is introduced, providing enhanced protection against discrepancies between plant and model parameters. Additionally, a realistic friction model is incorporated to simulate dynamic frictional effects. The paper is organized into seven sections. Section two covers human lower extremity anatomy, which is crucial for anthropomorphic modeling of the exoskeleton robot. Section three describes the development of kinematic and dynamic models and friction modeling. Section four details the Realistic Model

Reference Computed Torque control architecture and controller stability analysis. Section five presents the simulation results along with a discussion, while section six verifies the robustness of the developed controller concerning modeling discrepancies. Finally, section eight concludes the research with summary remarks.

## 2. Human Lower Extremity Anatomy

The human lower extremity is a complex structure comprising bones, joints, and major muscles that work together to facilitate movement and support the body. The primary bones include the femur (thigh bone), patella (kneecap), tibia (shin bone), and fibula. The hip joint connects the femur to the pelvis and allows for a wide range of motion in multiple planes, providing approximately three degrees of freedom (flexion/extension, abduction/adduction, and internal/external rotation). The knee joint, formed by the femur, tibia, and patella, enables flexion and extension and internal-external rotation, contributing two degrees of freedom. The ankle joint connects the tibia and fibula to the foot, allowing for dorsiflexion and plantarflexion, with additional slight movements of inversion and eversion.

Key muscles include the quadriceps, which extend the knee; the hamstrings, responsible for knee flexion; and the gastrocnemius and soleus muscles in the calf, which facilitate plantarflexion of the foot. The gluteal muscles, including the gluteus maximus, medius, and minimus, play vital roles in hip stabilization and movement. Together, this intricate anatomy supports weight-bearing, locomotion, and balance, allowing for efficient and coordinated movement during various activities. **Error! Reference source not found.** shows the human lower extremity degrees of freedoms.

### 2.1. Human Lower Extremity Ranges of Motion

The range of motion in the human lower extremity is essential for determining the joint ranges of motion in exoskeleton robots. Most healthy individuals exhibit similar ranges of motion. However, for physical therapy patients, the acceptable range values differ based on each individual's health condition. Average joint ranges of motion have been documented from various sources[34], [35] , [36].

The human lower extremity exhibits a wide range of motion across various joints, essential for mobility and functionality. The hip joint allows approximately 120 degrees of flexion, 30 degrees of extension, 45 degrees of abduction, and 30 degrees of adduction, along with internal and external rotation (both are 45 degrees). The knee joint permits about 135 degrees of flexion and limited extension. The ankle joint facilitates dorsiflexion of around 20 degrees and plantarflexion of approximately 50 degrees, along with inversion/pronation and eversion/supination movements (35 degrees and 15 degrees). This extensive range of motion enables activities such as walking, running, jumping, and climbing, contributing to overall mobility and stability.

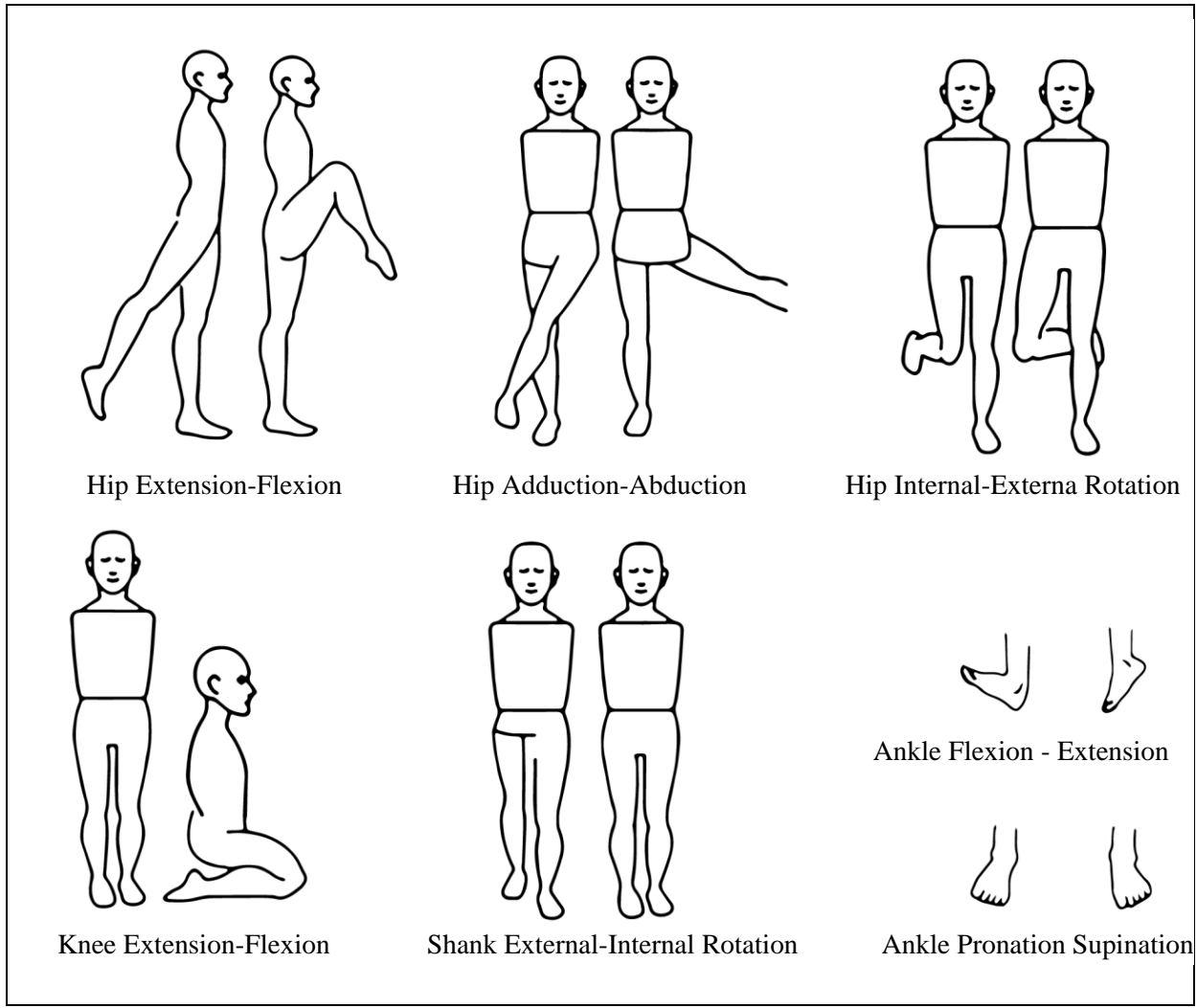

Figure 1 Human Lower extremities degrees of freedom

## 2.2. Human Lower Extremity Anthropometric Parameters

Anthropometric properties are essential for developing kinematic and dynamic models of the human lower extremity. Nikolova and Toshev conducted a study analyzing these properties using data from 5,290 individuals, including 2,435 males and 2,855 females, with their findings published in 2008 [37]. Additionally, Contini created empirical equations to determine anthropometric parameters of the human upper and lower extremity based on the subject's weight and height. For this article, we used the empirical formulas proposed by Contini. Equations (1) to (21) outline the calculations used for determining the anthropometric parameters necessary for dynamic modeling and simulation.

$$B_d = 0.6905 + 0.0297 C \frac{lb}{ft^3}, where\ C = HW^{-\frac{1}{3}}, \tag{1}$$

In equation (1) $B_d$ is the body density ($lbs/ft^3$), $H$ is the height of the subject ($inch$), $W$ is the body weight of the subject ($lbs$). The thigh, shank and foot densities are determined by the Equations (2) to (4),

$$T_d = 1.035 + 0.814 * B_d \frac{lb}{ft^3} \tag{2}$$

$$S_d = 1.065 + B_d \frac{lb}{ft^3} \tag{3}$$

$$F_d = 1.071 + B_d \frac{lb}{ft^3} \tag{4}$$

In Equations. (2)-(4), $T_d$ is the thigh density, $S_d$ is the shank density and $F_d$ is the foot density. The whole-body volume ($B_v$) was calculated using body weight and body density.

$$B_v = \frac{W}{B_d}\ ft^3 \tag{5}$$

The volume of thigh, shank, and foot were calculated as follows:

$$T_v = 0.0922 * B_v\ ft^3 \tag{6}$$

$$S_V = 0.0464 * B_v\ ft^3 \tag{7}$$

$$F_v = 0.0124 * B_v\ ft^3 \tag{8}$$

Approximate weights of the thigh, shank, and foot were calculated as given using Equations (9)-(11)

$$T_m = T_v * T_d\ lbs \tag{9}$$

$$S_m = S_v * S_d \; lbs \tag{10}$$

$$F_m = F_v * F_d \; lbs \tag{11}$$

The length of the thigh ($T_l$), shank ($S_l$), foot ($F_l$) and ankle to lower face of the foot ($A_g$) were calculated as shown in Equations (12)-(15),

$$T_l = 0.245 * H \; inch, \tag{12}$$

$$S_l = 0.285 * H \; inch, \tag{13}$$

$$F_l = 0.152 * H \; inch, \tag{14}$$

$$A_g = 0.043 * H \; inch \tag{15}$$

The locations of the center of the mass from the proximal joint (for thigh ($T_{cm}$), shank ($S_{cm}$), foot ($F_{cm}$)) are given by Equations (16)-(18),

$$T_{cm} = 0.41 * T_l \; inch, \tag{16}$$

$$S_{cm} = 0.393 * S_l \; inch, \tag{17}$$

$$F_{cm} = 0.445 * F_l \; inch \tag{18}$$

The empirical equations for the inertial properties of the thigh $T_i$, shank $S_i$, and foot $F_i$ are given in Equation (19) to Equation (21),

$$T_i = \begin{bmatrix} T_m(0.124 * T_l)^2 & 0 & 0 \\ 0 & T_m(0.267 * T_l)^2 & 0 \\ 0 & 0 & T_m(0.267 * T_l)^2 \end{bmatrix} \tag{19}$$

$$S_i = \begin{bmatrix} S_m(0.281 * S_l)^2 & 0 & 0 \\ 0 & S_m(0.114 * S_l)^2 & 0 \\ 0 & 0 & S_m(0.275 * S_l)^2 \end{bmatrix} \tag{20}$$

$$F_i = \begin{bmatrix} F_m(0.124 * F_l)^2 & 0 & 0 \\ 0 & F_m(0.245 * F_l)^2 & 0 \\ 0 & 0 & F_m(0.257 * F_l)^2 \end{bmatrix} \tag{21}$$

## 3. Human Lower Extremity Kinematic and Dynamic Modeling

The human lower extremity exoskeleton robot is designed to be closely attached to the human body, aiming to replicate human joints and movements. The distribution of moment of inertia in the exoskeleton mirrors that of the human lower limb. To maintain a general approach rather than focusing on a specific configuration, the anatomical structure and anthropometric parameters of the human lower limb were used into this simulation. We are actively developing an exoskeleton robot for lower extremity rehabilitation. Figure 2 illustrates the human lower extremity exoskeleton robot currently under development.

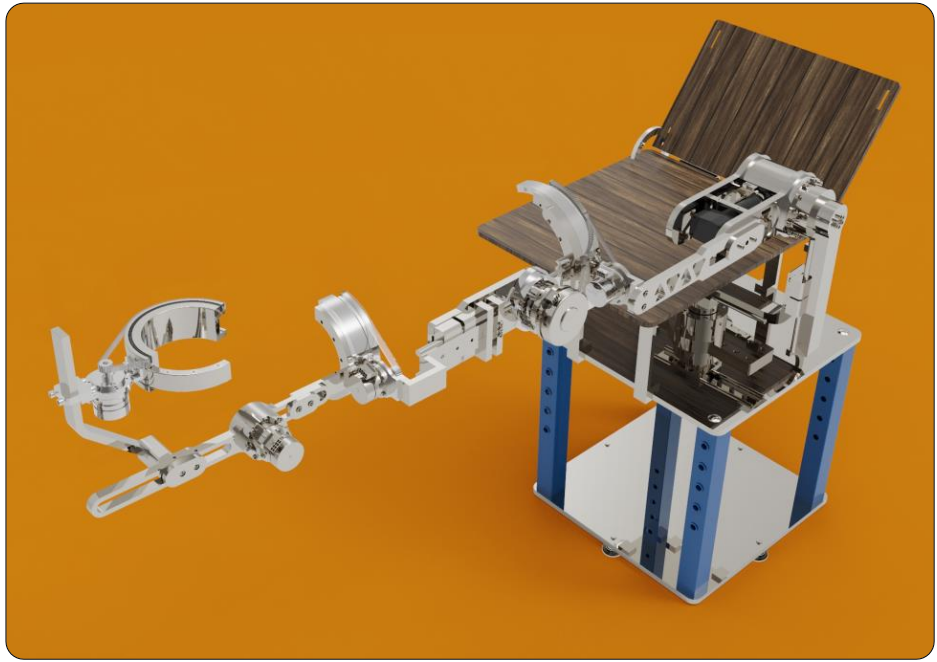

Figure 2 Human lower extremity rehabilitation exoskeleton robot

### 3.1. Kinematic and Dynamic Modeling

The kinematic model was developed using the modified Denavit-Hartenberg (DH) parameters method. A separate link frame was assigned to each degree of freedom. Figure 3 illustrates the link frame assignments according to the modified DH parameters. The modified D-H parameters for each joint are given in Table 2.

Table 2: Modified D-H parameters of human lower extremity

| Gesture name | Joint variable $\theta_i$ | Link offset $d_i$ | Link length $a_{i-1}$ | Link twist $\alpha_{i-1}$ |
|---|---|---|---|---|
| **Hip Abduction/Adduction** | $q_1$ | 0 | 0 | 0 |
| **Hip Flexion/Extension** | $q_2 - \frac{\pi}{2}$ | 0 | 0 | $-\frac{\pi}{2}$ |
| **Hip Internal/External rotation** | $q_3$ | $-l_1$ | 0 | $-\frac{\pi}{2}$ |
| **Knee Flexion/Extension** | $q_4$ | 0 | 0 | $\frac{\pi}{2}$ |
| **Knee Internal rotation** | $q_5$ | $-l_2$ | 0 | $-\frac{\pi}{2}$ |
| **Ankle Dorsiflexion/Plantarflexio** | $q_6 - \frac{\pi}{2}$ | 0 | 0 | $\frac{\pi}{2}$ |
| **Ankle Pronation/Supination** | $q_7$ | 0 | $a_1$ | $-\frac{\pi}{2}$ |

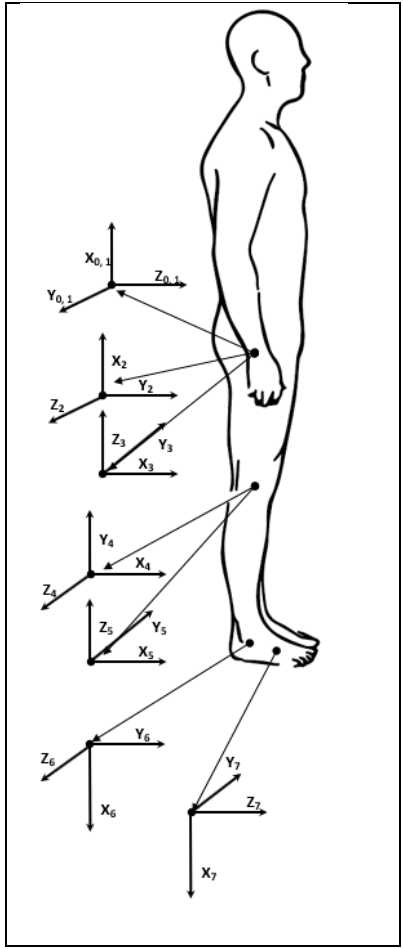

Figure 3 Link Frame Assignment

The general form of the transformation matrix is presented in Equation (22):

$$^{i-1}_{i}T = \begin{bmatrix} \cos\theta_i & -\sin\theta_i & 0 & \alpha_{i-1} \\ \sin\theta_i \cos\alpha_{i-1} & \cos\theta_i \cos\alpha_{i-1} & -\sin\alpha_{i-1} & -\sin\alpha_{i-1}\, d_i \\ \sin\theta_i \sin\alpha_{i-1} & \cos\theta_i \sin\alpha_{i-1} & \cos\alpha_{i-1} & \cos\alpha_{i-1} d_i \\ 0 & 0 & 0 & 1 \end{bmatrix} \quad (22)$$

By substituting the modified DH parameter values from Table 3 into Equation. (22), the resulting transformation matrices, Equation (23) to Equation (29), were obtained.

$$^{0}T_1 = \begin{bmatrix} \cos(\theta_1) & -\sin(\theta_1) & 0 & 0 \\ \sin(\theta_1) & \cos(\theta_1) & 0 & 0 \\ 0 & 0 & 1 & 0 \\ 0 & 0 & 0 & 1 \end{bmatrix} \quad (23)$$

$$^1T_2 = \begin{bmatrix} \sin(\theta_2) & \cos(\theta_2) & 0 & 0 \\ 0 & 0 & 0 & 0 \\ \cos(\theta_2) & -\sin(\theta_2) & 1 & 0 \\ 0 & 0 & 0 & 1 \end{bmatrix} \tag{24}$$

$$^2T_3 = \begin{bmatrix} \cos(\theta_3) & -\sin(\theta_3) & 0 & 0 \\ 0 & 0 & 0 & -l_1 \\ -\sin(\theta_3) & -\cos(\theta_3) & 1 & 0 \\ 0 & 0 & 0 & 1 \end{bmatrix} \tag{25}$$

$$^3T_4 = \begin{bmatrix} \cos(\theta_4) & -\sin(\theta_4) & 0 & 0 \\ 0 & 0 & -1 & 0 \\ \sin(\theta_4) & \cos(\theta_4) & 0 & 0 \\ 0 & 0 & 0 & 1 \end{bmatrix} \tag{26}$$

$$^4T_5 = \begin{bmatrix} \cos(\theta_5) & -\sin(\theta_5) & 0 & 0 \\ 0 & 0 & 1 & -l_2 \\ -\sin(\theta_5) & -\cos(\theta_5) & 0 & 0 \\ 0 & 0 & 0 & 1 \end{bmatrix} \tag{27}$$

$$^5T_6 = \begin{bmatrix} \sin(\theta_6) & \cos(\theta_6) & 0 & 0 \\ 0 & 0 & -1 & 0 \\ -\cos(\theta_6) & \sin(\theta_6) & 0 & 0 \\ 0 & 0 & 0 & 1 \end{bmatrix} \tag{28}$$

$$^6T_7 = \begin{bmatrix} \cos(\theta_7) & -\sin(\theta_7) & 0 & a_1 \\ 0 & 0 & 1 & 0 \\ -\sin(\theta_7) & -\cos(\theta_7) & 0 & 0 \\ 0 & 0 & 0 & 1 \end{bmatrix} \tag{29}$$

The homogeneous transformation matrix determines the position and orientation of the end frame relative to the base frame. The homogeneous transformation matrix was constructed by multiplying the individual transformation matrices from Equation (23) to Equation (29). For the lower extremity, the homogeneous transformation matrix represents the position and orientation of the foot relative to the hip joint.

$$^0_7T = \begin{bmatrix} ^0_1T \, ^1_2T \, ^2_3T \, ^3_4T \, ^4_5T \, ^5_6T \, ^6_7T \end{bmatrix} \tag{30}$$

Lagrange's method is a widely used mathematical approach for deriving dynamic equations of motion. It involves differentiating energy terms with respect to the system variables and time. The dynamic model of the human lower extremity was developed using Lagrange's method [38]. The kinetic energy of a link at any given moment can be calculated using Equation (31).

$$k_i = \left[\frac{1}{2} m_i . v_{ci}^T . v_{ci} + \frac{1}{2}\, ^i\omega_i^T . ^{ci}I_i . \omega_i\right] \tag{31}$$

In Equation (31), $m_i$ represents the mass of the link, while $v_{ci}$ and $\omega_{ci}$ denote the linear and angular velocities of the link at its center of mass. $I_i$ is the moment of inertia of the link about its center of mass. The total kinetic energy of the model can be calculated using Equation (32).

$$k = \sum_{i=1}^{n} k_i \tag{32}$$

In Equation, (32), $n$ represents the total number of degrees of freedom, $n = 7$ for the developed human lower extremity dynamic model.

The potential energy of any link can be calculated using Equation (33),

$$u_i = -m_i . \,^0g^T . \,^0P_{ci} + u_{ref} \tag{33}$$

$$u = \sum_{i=1}^{n} u_i \tag{34}$$

In Equation (33), $m_i$ represents the mass of the link, $g$ is the gravitational acceleration, and $^0P_{ci}$ denotes the position of the link's center of mass relative to the ground reference. Finally, by using the Lagrange energy method, the required joint torque can be determined as follows:

$$\tau_i = \left[\frac{d}{dt}\frac{\delta k}{\delta \dot{\theta}_i} - \frac{\delta k}{\delta \theta_i} + \frac{\delta u}{\delta \theta_i}\right] \tag{35}$$

The dynamic equations of motion for the robot are expressed in Eq. (36),

$$\tau_{Joint} = [M(\theta)\ddot{\theta} + V(\theta,\dot{\theta}) + G(\theta)] \tag{36}$$

In Equation (36), $M(\theta)$ is the mass matrix, a $(7x7)$ symmetric positive definite matrix, $V(\theta,\dot{\theta})$ is $(7x1)$ matrix represents the Coriolis and the centripetal terms, and $G(\theta)$ is $(7x1)$ matrix representing the gravitational term. $\tau_{Joint}$, $(7x1)$ matrix, represents the joint torque requirements. The robot's dynamic equation of motion can be rewrite as:

$$\ddot{\theta} = \left[M(\theta)^{-1}\left(\tau_{Joint} - V(\theta,\dot{\theta}) - G(\theta)\right)\right] \quad (37)$$

Since $M(\theta)$ is a positive definite matrix, $M(\theta)^{-1}$ always exists. Figure 4 shows a schematic diagram of the ideal/model robot dynamics without counting the joint frictions.

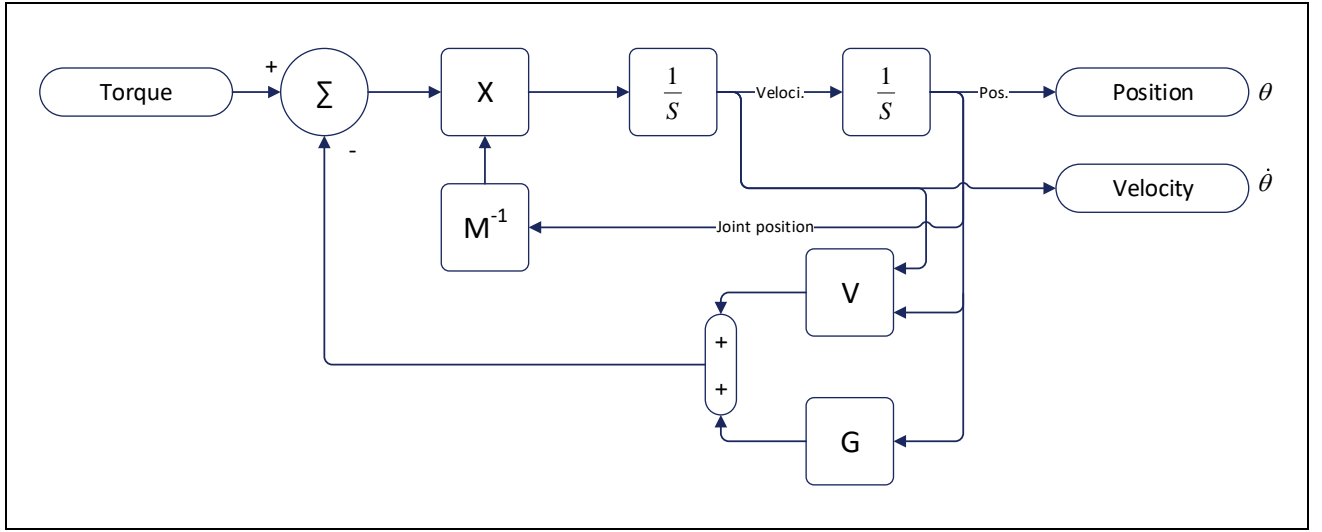

Figure 4 Internal architecture of the robot model

Friction between two mating parts with relative motion is unavoidable. After considering joint friction torques, the robot dynamical equations become,

$$\tau_{Joint} = \left[M(\theta)\ddot{\theta} + V(\theta,\dot{\theta}) + G(\theta) + \tau_{friction}\right] \quad (38)$$

where,

$$\tau_{friction} = \left[F(\dot{\theta})\right] \quad (39)$$

Equation (38) can also be written as,

$$\ddot{\theta} = \left[M(\theta)^{-1}\left(\tau_{Joint} - V(\theta,\dot{\theta}) - G(\theta) - F(\dot{\theta})\right)\right] \quad (40)$$

Figure 5 presents the schematic diagram of the robot dynamics with friction effects.

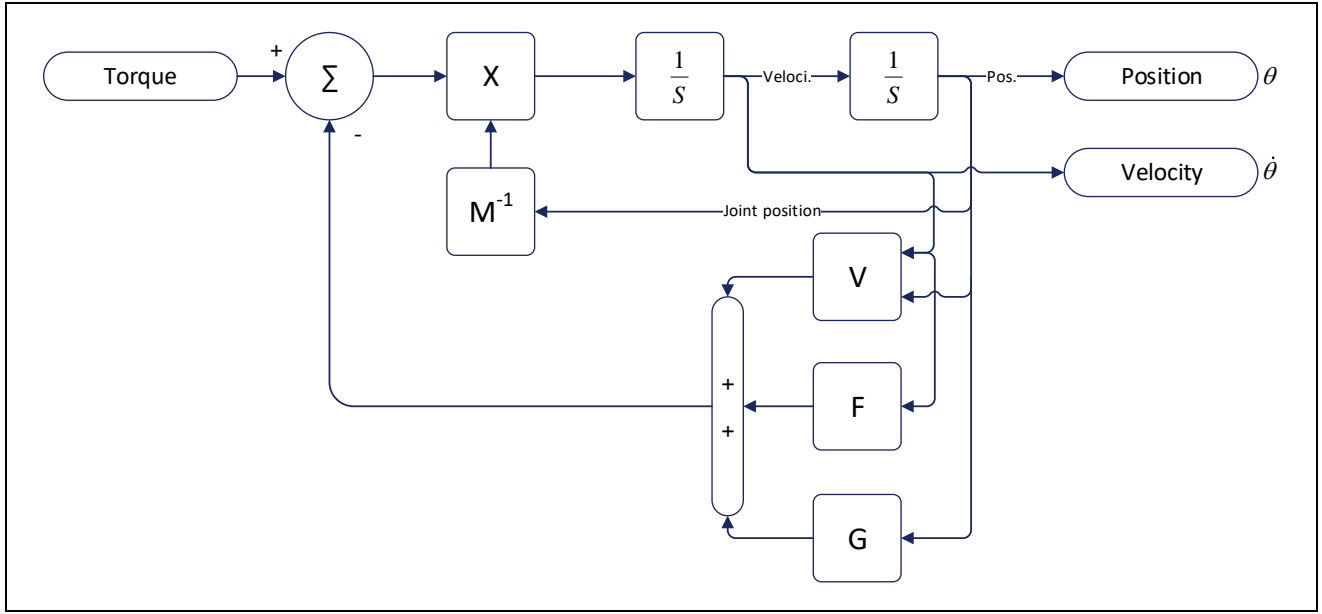

Figure 5 Internal architecture of the physical Robot model

## 3.2. Friction Modeling

In robotic manipulators, links are connected by bearings, transmissions, or seals, where relative motion at the joints leads to the generation of friction forces. These friction forces can account for as much as 50% of the transmitted force or torque [39]. A robust control system is essential for compensating for the effects of friction. The magnitude of friction force or torque depends on various factors, such as surface roughness, lubricant viscosity, transmitted load, temperature, and relative velocity between contact surfaces. Since these parameters are variable, it is challenging to develop a comprehensive theoretical friction model.

Most friction models are empirical and have been successfully used for years. Researchers have described friction through models like the Coulomb friction model, Viscous friction model, Stribeck effect, pre-sliding behavior, stiction phase, and hysteresis effects. Advanced models like the Dahl, LuGre, and Karnopp models [40] also exist. In this paper, a friction model combining the Coulomb, Viscous, and Stribeck effects, equivalent to the LuGre model [41], is used. The next section will explain the formulation of the friction model in detail.

**Coulomb friction** ($T_C$) is characterized by a constant friction torque regardless of the velocity or movement of the system, meaning that it remains uniform at any given time. In contrast, **viscous friction** ($T_V$) produces a resistive torque that is directly proportional to the relative velocity between the surfaces in contact. **Stribeck friction** ($T_S$) describes a more complex behavior observed at low velocities, where there is a noticeable decrease in friction force, resulting in a negative slope in the friction-velocity relationship. Together, these friction models describe the various types of resistive forces that can occur in mechanical systems.

Equation (41) presents the friction model, with Equations (41) to (43) were used to calculate the joint friction torque:

$$T = \sqrt{(2e)}(T_{brk} - T_C) \cdot \exp\left(-\left(\frac{\omega}{\omega_{St}}\right)^2\right) \cdot \frac{\omega}{\omega_{St}} + T_C \cdot \tanh\left(\frac{\omega}{\omega_{Coul}}\right) + f\omega \qquad (41)$$

$$\omega_{St} = \omega_{brk}\sqrt{2} \qquad (42)$$

$$\omega_{Coul} = \frac{\omega_{brk}}{10} \qquad (43)$$

where,

$T$ is the total friction torque

$T_C$ is the Coulomb friction torque

$T_{brk}$ is the breakaway friction torque: This is the sum of the Coulomb and Stribeck friction torques near zero velocity, often referred to as breakaway friction.

$\omega_{brk}$ is the breakaway friction velocity: The velocity at which Stribeck friction reaches its peak, and the sum of the Stribeck and Coulomb friction equals the breakaway friction torque.

$\omega_{St}$ is the Stribeck velocity threshold

$\omega_{Coul}$ is the Coulomb velocity threshold

$\omega$ is the input angular velocity

$f$ is the viscous friction coefficient: A proportionality constant between the friction torque and angular velocity, which must have a positive value.

Figure 6 and Figure 7 presents the relation between the angular velocity and the generated friction torque.

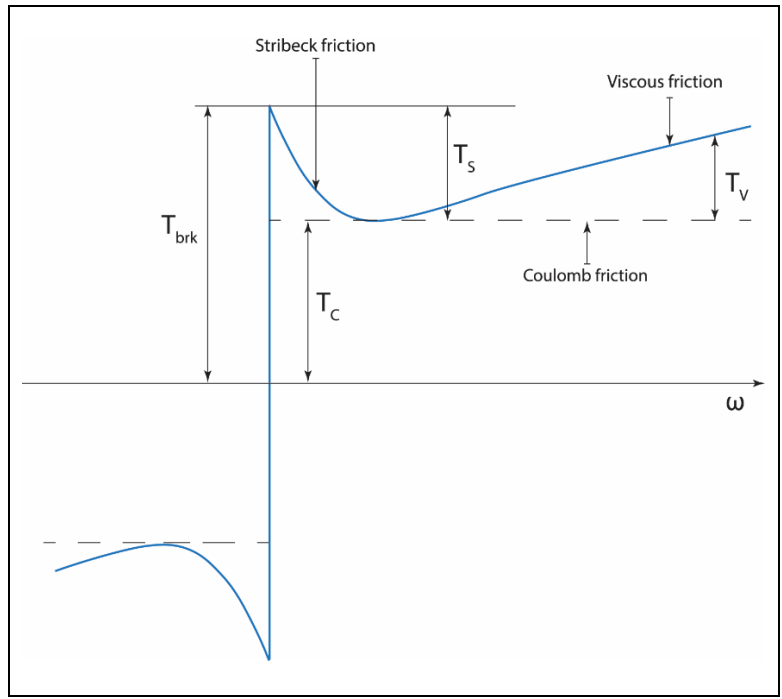

Figure 6 Friction model (Combined Coulomb, viscus and Stribeck effects)

Figure 7 shows the simulated friction torque based on Equations (41) to (43)

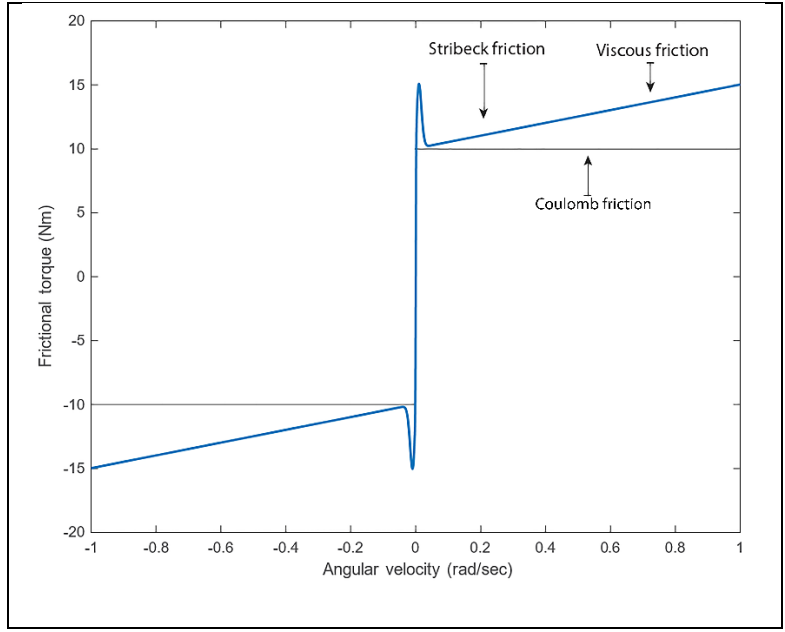

Figure 7 Simulation of the friction model

In the above simulation, the following assumptions were made:

$T_{Peak} = 100\ Nm, \omega = 100\ rad/sec, f = 5\ Nm/(rad/sec), T_C = 0.1 * T_{Peak}\ Nm,$

$\omega_{brk} = 0.01\ rad/sec, T_{brk} = 0.15 * T_{Peak}\ Nm$

Friction between two interacting parts with relative motion is unavoidable. By considering the joint friction torques, the robot's dynamics are modified as follows:

$$\tau_{Joint} = \left[M(\theta)\ddot{\theta} + V(\theta,\dot{\theta}) + G(\theta) + \tau_{friction}\right] \qquad (44)$$

where,

$$\tau_{friction} = F(\dot{\theta}) \qquad (45)$$

Equation (38) can be written in the form of Equation (46)

$$\ddot{\theta} = M(\theta)^{-1}\left[\tau - V(\theta,\dot{\theta}) - G(\theta) - F(\dot{\theta})\right] \qquad (46)$$

The next section (section 4) will explain the construction of the Realistic Model Reference Computed Torque Controller (RMRCTC).

# 4. Realistic Model Reference Computed Torque Control

The realistic model reference computed torque controller is an enhanced version of the standard model reference computed torque controller. One of the key limitations of the traditional model reference computed torque controller is that it requires the dynamic model of the plant to be executed twice during each sampling period, demanding significant computational power [42]. The computed torque control scheme, theoretically based on the plant's inverse dynamics, is utilized within the model reference computed torque controller for real-time torque estimation. The following section will discuss the computed torque controller, the model reference computed torque controller, and the realistic model reference computed torque controller in detail.

## 4.1. Computed Torque Controller:

The computed torque controller is widely recognized as one of the most effective model-based control methods, leveraging inverse dynamics. It consists of two key components: a linearization mechanism and a control action. However, its primary limitation is the need for an accurate dynamic model of the system, making it more suitable for applications such as industrial robotics, where model parameters are well-defined. In contrast, rehabilitation exoskeletons must adapt to users with varying body sizes and shapes, making it impractical to accurately measure the mass and inertia of individual limbs. Since the controller's stability and trajectory tracking performance rely heavily on precise modeling, this presents a significant challenge. Furthermore, the entire dynamic model must be recalculated during each sampling period, placing a heavy computational burden on the control system..

It performs two key functions: first, it linearizes the nonlinear system using feedback linearization, and second, it generates a control input that meets performance criteria [38]. Figure 8 illustrates the architecture of the CTC scheme.

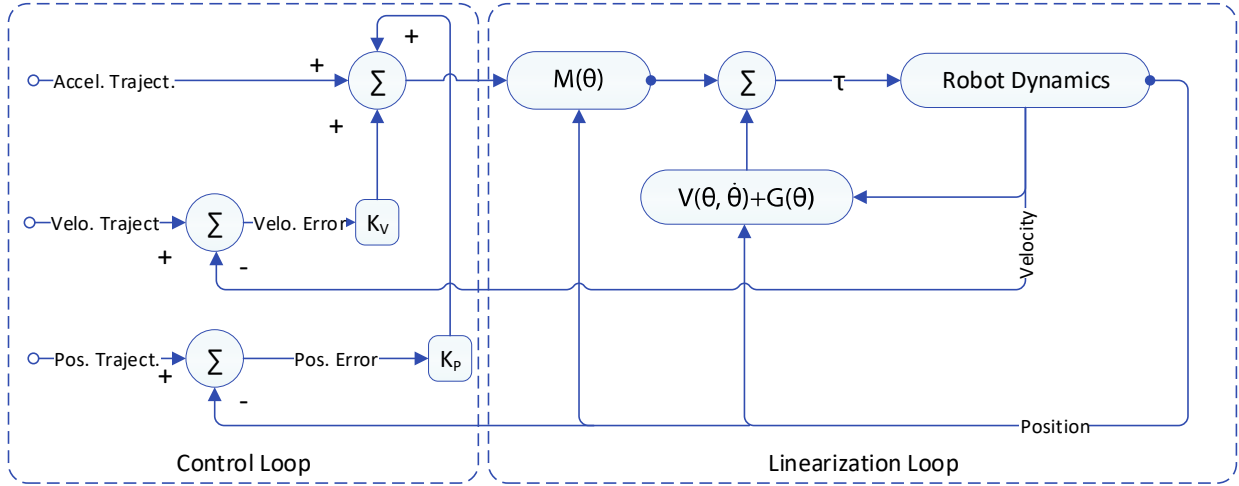

Figure 8 Schematic diagram of the computed torque controller

In Figure 8, $M(\theta) \in \mathbb{R}^{n \times n}$ represents the robot mass matrix, $V(\theta, \dot{\theta}) \in \mathbb{R}^{n \times 1}$ indicates the Coriolis and centrifugal forces, and $G(\theta) \in \mathbb{R}^{n \times 1}$ denotes the gravitational force. Figure 8 clearly shows that the CTC relies on the robot's dynamic model for both linearizing the robot dynamics and generating the control input. Consequently, any discrepancies between the robot and its model can result in instability and uncontrolled behavior, making it essential to protect the system from such discrepancies.

## 4.2. Model Reference Computed Torque Controller

A promising alternative to the computed torque controller is the model reference computed torque controller. In this approach, the plant's model is executed at every sampling time with the help of a computed torque controller for estimating the torque required to track the provided trajectories. The predicted torque is then applied to the physical system. The trajectory tracking error between the model and the physical plant is measured, and any discrepancies are corrected using an additional PID controller. Figure 9 shows the schematic diagram of a model reference computed torque controller. In essence, the model reference computed torque control scheme requires the robot's dynamics to be executed twice in each sampling period (Robot's dynamic model and M, G, V matrices). This significantly increases the computational load on the digital controlle. High speed multiple core digital computer is required to run the control algorithms. The control system consumes a significant amount of energy to run the control algorithms.

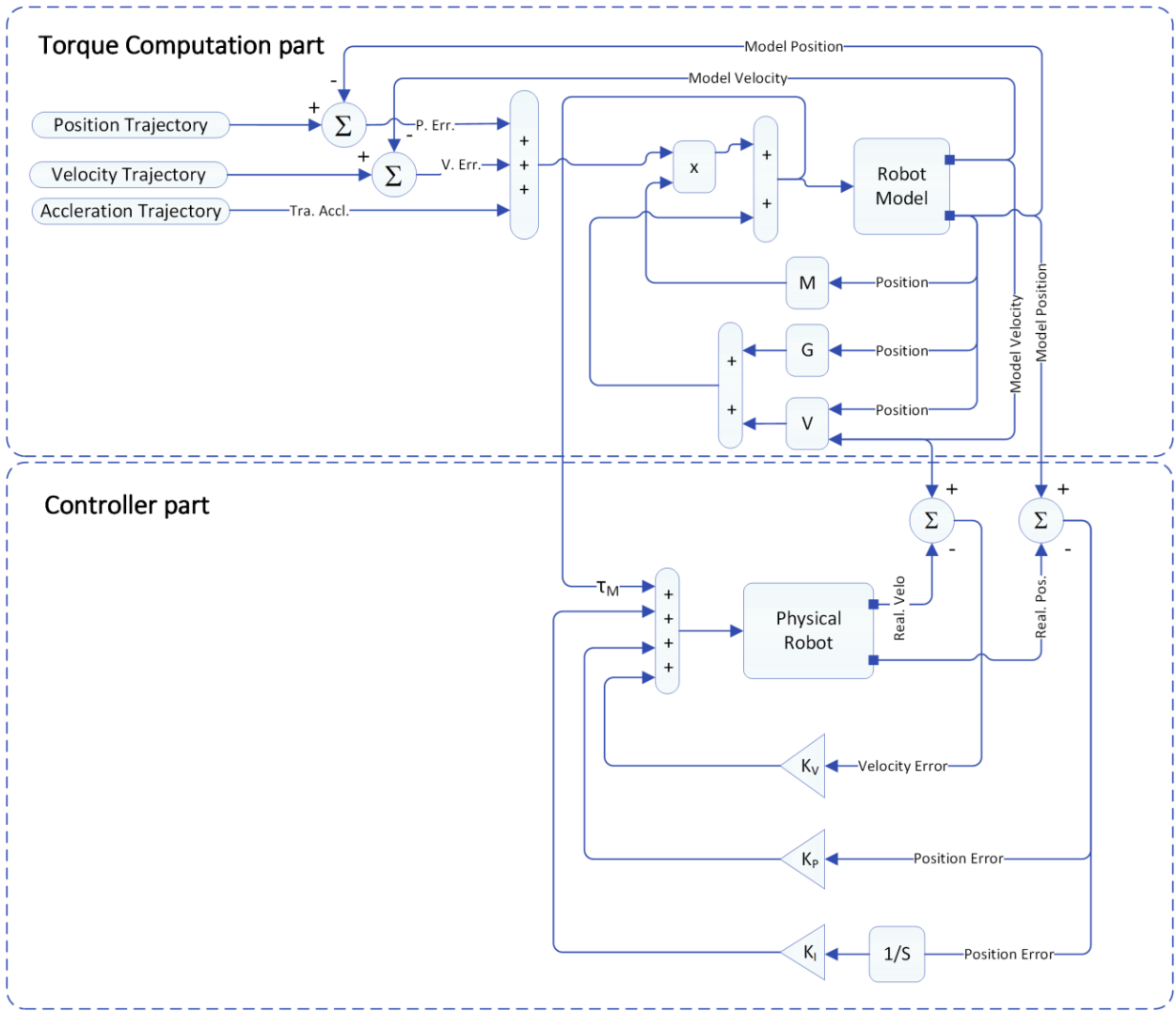

Figure 9 Architecture of Model Reference Computed Torque Control

## 4.3. Realistic Model Reference Computed Torque Controller

In order to make the Model Reference Computed Torque Controller more efficient and realistic multiple changes were made in the Realistic Model Reference Computed Torque Controller.

Equation (36) shows that the total torque required to operate the robot is used for accelerating the robot's links and payloads (M matrix), compensating for gravitational effects (G matrix), and counteracting Coriolis and centrifugal effects (V matrix). Often joints friction are considered as the disturbances. It has been observed that the magnitude of Coriolis and centrifugal forces is significantly smaller compared to the torque needed for acceleration and gravity compensation.

Calculating the Coriolis and centrifugal terms at every sampling period for the robot's specific position and velocity demands more computational power than executing the mass and gravity matrices (M and G matrices).

To optimize the realistic model reference computed torque controller, we excluded the Coriolis and centrifugal terms from the dynamic model, treating them as disturbances. Additionally, we divided the control algorithm into two loops: slower loop for joint torque estimation and faster loop contains the PID controller for minimizing the tracking error between the reference trajectory and the plant output. The slower loop run at (100Hz), while the faster loop run at 1kHz. Running 2 loops at two different rates reduces the computational load and energy consumption for real time computing. Predicting torque at slower speed cause some inaccuracies. The PID controller was used to remove the errors and improve the robustness.The developed realistic model reference computed torque controller showed excellent trajectory tracking performance, maintaining small tracking errors even in the presence of disturbances. By excluding the Coriolis and centrifugal force terms $V(\theta, \dot{\theta})$, it achieves significantly higher energy efficiency compared to traditional computed torque control and model reference computed torque control methods. Figure 11 shows the schematic diagram of the realistic model reference computed toque controller.

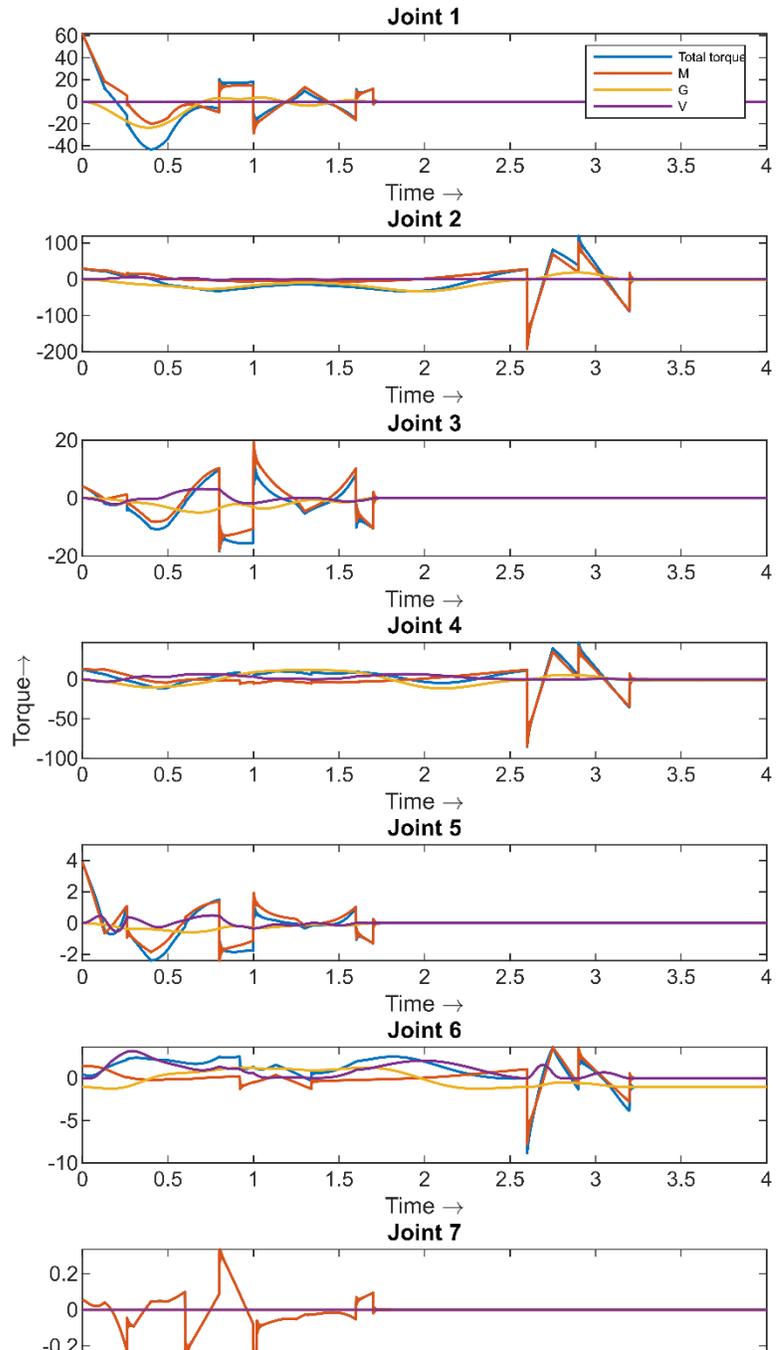

Figure 10 Comparison between the total torque, torque required for link acceleration (M), to overcome the gravity (G), Coriolis and centrifugal forces (V).

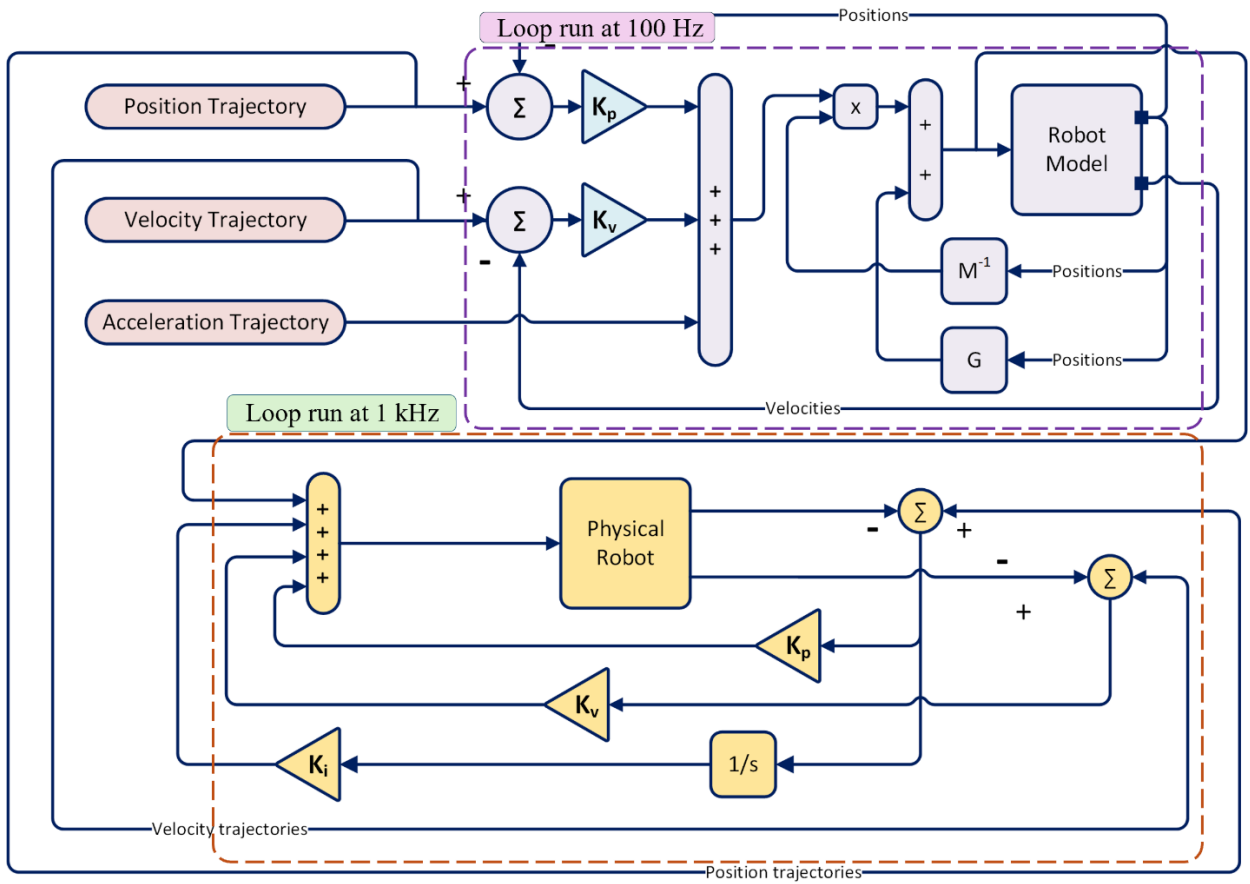

Figure 11 Realistic Model Reference Computed Torque Control Architecture

**Stability Analysis of the Proposed Realistic Model Reference Computed Torque Controller:**

Ensuring controller stability is a very fundamental requirement of a control system. The following section will prove the combined system stability by computing the stability of the individual loops.

**Loop1:**

The dynamic model of the robot can be represented by Equation (47),

$$[M(\theta)\ddot{\theta} + V(\theta,\dot{\theta}) + G(\theta)] = \tau_{Joint} \tag{47}$$

The control torque for the model is calculated by using Equation (48),

$$\tau_M = M(\theta)[\ddot{\theta}_d + K_v(\dot{\theta}_d - \dot{\theta}) + K_P(\theta_d - \theta)] \tag{48}$$

By comparing equations (47) and (48),

$$(\ddot{\theta}_d - \ddot{\theta}) + K_V(\dot{\theta}_d - \dot{\theta}) + K_P(\theta_d - \theta) = 0 \tag{49}$$

The robot tracking error is defined as $E = (\theta_d - \theta)$. Using equation (49), the error dynamics can be expressed as:

$$\ddot{E} + K_V \dot{E} + K_P E = 0 \tag{50}$$

By applying the Laplace transform on Eqn. (50), the closed-loop characteristic polynomial can be written as,

$$\Delta_C(S) = |S^2 I + K_V S + K_P| \tag{51}$$

where,
$$K_P = diag\{K_{Pi}\} \tag{52}$$

$$K_V = diag\{K_{Vi}\} \tag{53}$$

According to the Routh-Hurwitz stability criteria in

Table 3, equation (51) is asymptotically stable as long as the gain matrices $(K_P, K_V)$ are positive definite.

Table 3 Routh Hurtz stability criteria,

$$\begin{array}{c|cc} S^2 & 1 & K_P \\ S^1 & K_V & 0 \\ S^0 & K_P & \end{array}$$

**Loop 2:**

Calculates the deviation of the plant output from the model $(e, \dot{e})$ and forces the plant to compensate for the tracking errors in a controlled way. Based on the developed RMRCTC scheme, the plant input torque $(\tau_P)$ can be calculated as,

$$\tau_P = \tau_M + K_I \int (\theta_M - \theta_P) + K_P(\theta_M - \theta_P) + K_V(\dot{\theta}_M - \dot{\theta}_P) + \tau_f \tag{54}$$

In developing this controller, it is assumed that the model and the plant share the same dynamical structure, differing only in their parameters.

The tracking error $(E, \dot{E})$ between the model and the plant is defined as,

$$E = (\theta_M - \theta_P) \tag{55}$$

$$\dot{E} = (\dot{\theta}_M - \dot{\theta}_P) \tag{56}$$

$$\tau_P = \tau_M + K_V \dot{E} + K_I \int E + K_P E + \tau_f \tag{57}$$

$$\tau_P(S) = \tau_M(S) + \frac{K_I E(S)}{S} + K_P E(S) + K_V S E(S) + \tau_f(S) \tag{58}$$

$$\tau_P(S) - \tau_M(S) - \tau_f(S) = \frac{(S^2 K_V + S K_P + K_I) E(S)}{S} \tag{59}$$

Using equation (59), the error dynamics can be expressed as:

$$\frac{E(S)}{\tau_P(S) - \tau_M(S) - \tau_f(S)} = \frac{S}{S^2 K_V + S K_P + K_I} \tag{60}$$

Equation (60) represents the transfer function of the tracking error

Table 4 shows that the control system will be asymptotically stable if the gain matrices $K_P, K_V, K_I$ are positive definite. Here, $K_P, K_I, K_V$ are the $(7x7)$ diagonal matrices.

$$K_P = diag\{K_P\} \tag{61}$$

$$K_I = diag\{K_I\} \tag{62}$$

$$K_V = diag\{K_V\} \tag{63}$$

Table 4: Routh-Hurwitz Stability Criteria for the Error Dynamics,

| | | |
|---|---|---|
| $S^2$ | $K_V$ | $K_I$ |
| $S^1$ | $K_P$ | 0 |
| $S^0$ | $K_I$ | |

An analysis of the stability of Loop1 and Loop 2, quickly leads to the conclusion that the proposed control system is asymptotically stable as long as the gain matrices are positive definite.

## 5. Simulation Results and Discussion

The trajectory tracking simulation was conducted in a MATLAB-Simulink® environment. The mass and inertial properties of the human lower extremities used in the simulation are listed in Table 5. Controller gains for both the computed torque control (Loop 1) and the PID controller (Loop 2) are also provided in Table 5. The simulation was performed for both sequential and simultaneous joint movements. Figure 12 to Figure 19 display the results of the developed controller for the human lower extremity's sequential and simultaneous joint movements.

Table 5 Simulation parameters for the HLE dynamic control using MRCTC

| | | | | | | |
|---|---|---|---|---|---|---|
| Subject mass | 163 lb (73.95 kg) | Distance between proximal joint and the center of | Thigh | 6.69 in (170 cm) | | |
| | | | Shank | 7.48 in (18.92 cm) | | |
| | | | Foot | 4.5 in (11.5 cm) | | |
| Subject height | 67 in (170.18 cm) | Thigh inertia $g.cm^2$ $(kg.m^2)$ | | $151*10^3$ (0.0151) | 0 | 0 |
| | | | | 0 | $700*10^3$ (0.070) | 0 |
| Thigh Mass | 12.45 lb (5.65 kg) | | | 0 | 0 | $700*10^3$ (0.070) |
| Shank mass | 7.67 lb (3.48 kg) | Shank inertia $g.cm^2$ $(kg.m^2)$ | | $648*10^3$ (0.06480) | 0 | 0 |
| | | | | 0 | $107*10^3$ (0.0107) | 0 |
| Foot Mass | 2.05 lb (0.93 kg) | | | 0 | 0 | $620*10^3$ (0.0620) |
| | | Foot inertia $g\,cm^2$ $(kg.m^2)$ | | $10*10^3$ (0.001) | 0 | 0 |
| Thigh-length | 16.14 in (41 cm) | | | 0 | $37*10^3$ (0.0037) | 0 |
| | | | | 0 | 0 | $41*10^3$ (0.0041) |
| Shank length | 18.89 in (48.79 cm) | PD controller gains Loop 1 | | PID controller gains Loop 2 | | |
| | | $K_P$ | [500, 500, 500, 500, 500, 500, 500] | $K_P$ | [$10^4$, $10^4$, $10^4$, $10^4$, $10^4$, $10^4$, $10^4$] | |
| | | | | $K_I$ | [250, 250, 250, 250, 250, 250, 250] | |
| Foot length | 10.23 in (25.88 cm) | $K_V$ | [7500, 7500, 7500, 7500, 7500, 7500, 7500] | $K_V$ | [55x$10^3$, 50x$10^3$, 55 x$10^2$, 3x$10^2$, 55x$10^2$, 55x$10^2$, 55x$10^2$] | |

The parameters used for modeling dynamic friction are listed below,

$$f = 0.1\ Nm/(rad/sec),\ \tau_{coulomb} = 0.1 * \tau_{peak}\ Nm,\ \tau_{brake} = 0.15 * \tau_{peak}\ Nm,$$

$$\omega_{stribeck} = \sqrt{2}\omega_{brake}\ rad/sec,\ \omega_{coulomb} = 0.10 * \omega_{brake}\ rad/sec$$

The peak torques were calculated based on the torque required to track the same trajectory using the dynamic model of the human lower extremity.

Figure 12 to Figure 15 illustrate the tracking performance, including trajectory tracking, tracking error, joint torque, and friction torque generated during sequential joint movement.

Figure 12 shows the input trajectory (Trajec. Joint #) supplied to the robot and the resultant output (Output Joint #) trajectories. The simulation result indicates the output closely follows the input trajectories with minimal error. Figure 13 provides a more detailed view of the tracking errors.

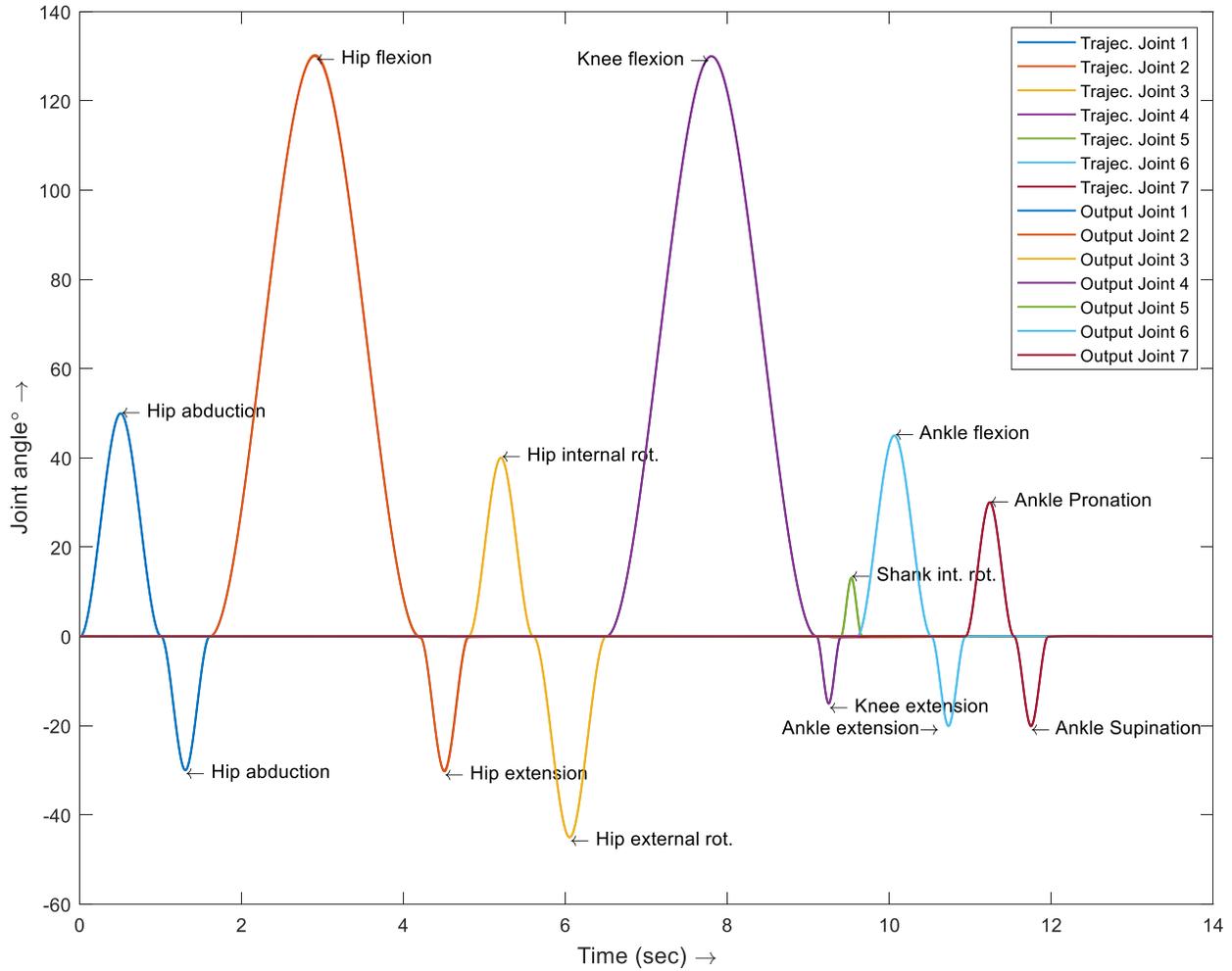

Figure 12 Simulation result using MRCTC for sequential joint movement

Figure 13 shows that the maximum tracking errors for sequential joint movements hip abduction-adduction, hip flexion-extension, hip internal-external rotation, knee flexion-extension, shank internal rotation, and ankle flexion-extension and pronation-supination were [0.38°, 0.68°, 0.12°, 0.29°, -0.19°, 0.17°, 0.16°] respectively.

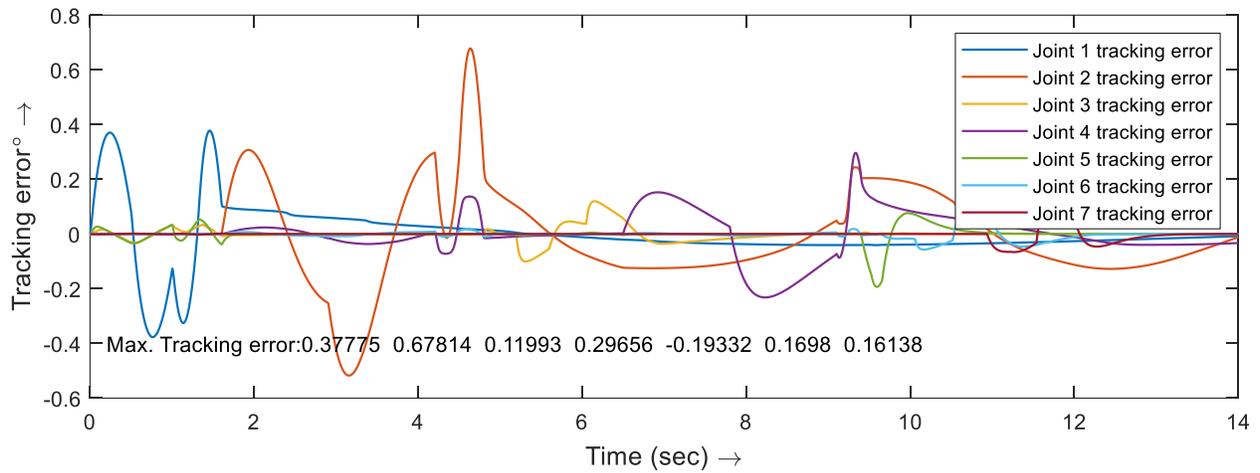

Figure 13 Tracking error using MRCTC for sequential joint movement

Figure 14 shows the joint torque needed for the model (Orange color) and the physical robot joints (blue) during the tracking of sequential joint movements. It is observed that the plant requires more torque than the model. The torque required by the plant consists of two components: firstly, the torque necessary for the model for tracking the trajectory and secondly the additional torque required for compensating the differences between the plant input trajectory and the plant's output.

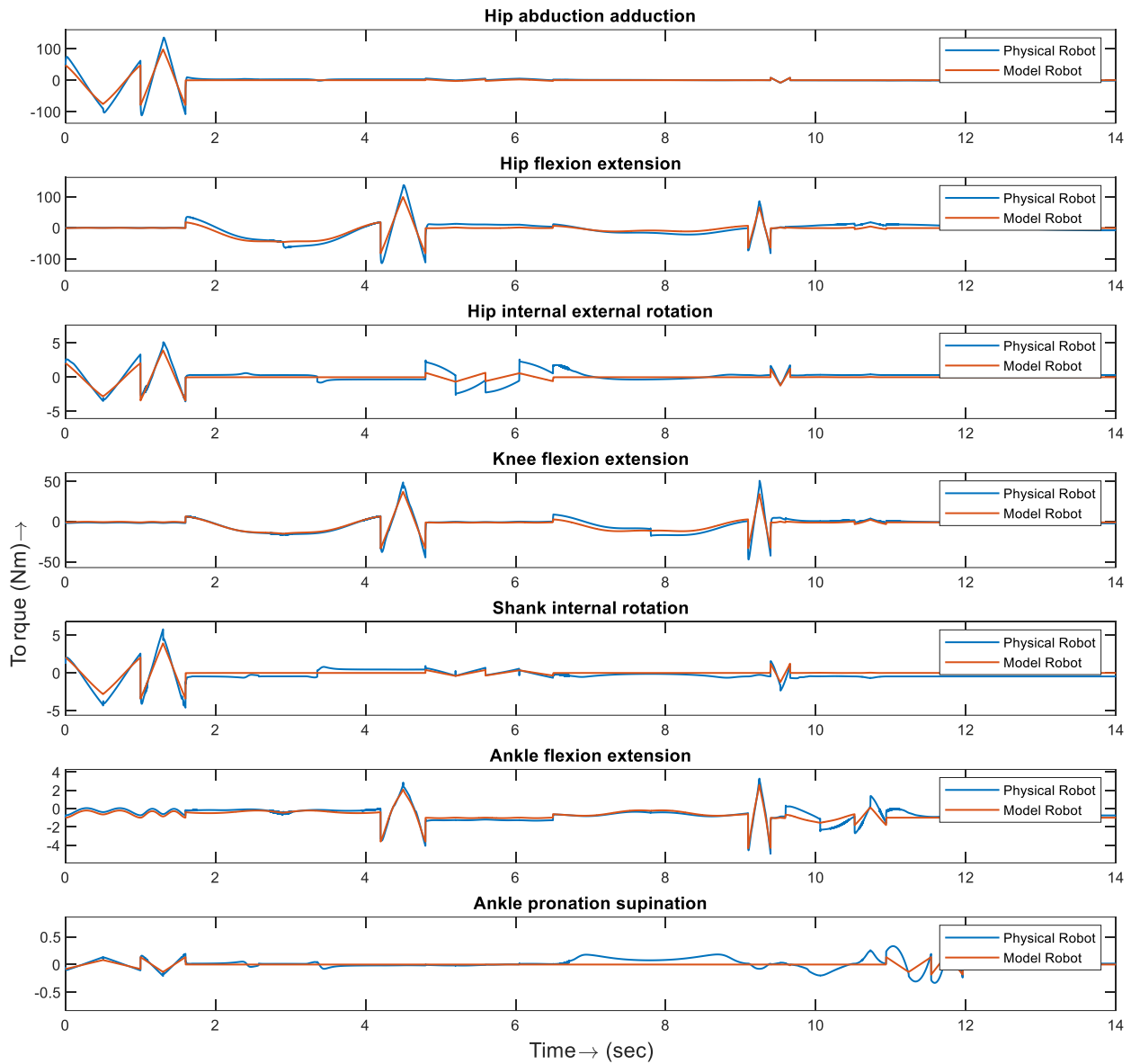

Figure 14 Joint torque required during sequential joint movement using MRCTC

Figure 15 presents the frictional torque generated in the joints during sequential joint movements. Equation (41) was used for estimating the dynamic friction torque generated in the joints due to relative motion between the mating surfaces. The simulation results indicate that joint 2 experienced the highest friction, followed by joints 1 and 4. In comparison, joints 3, 5, 6, and 7 produced significantly lower friction torque than joints 1, 2, and 4.

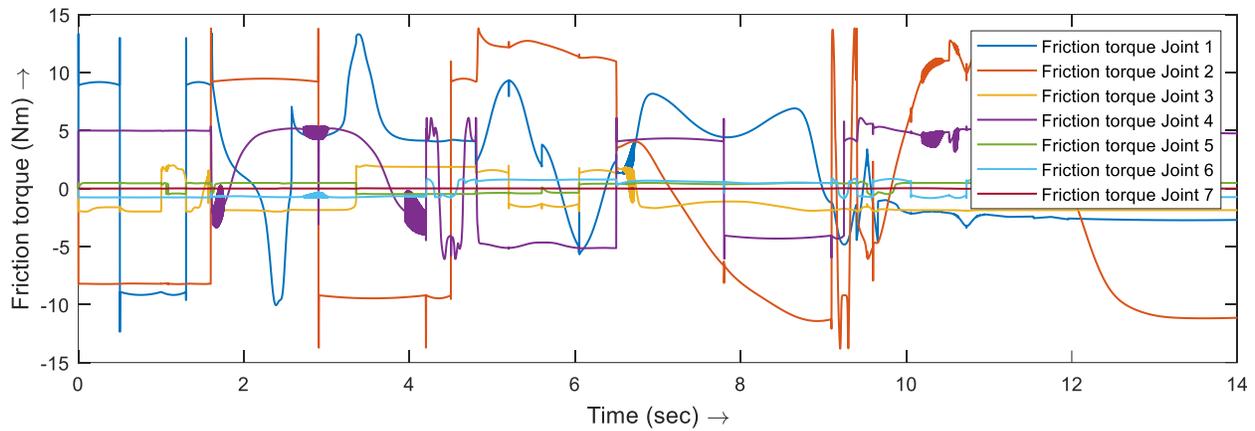

Figure 15 Joint friction torque required during sequential joint movement

Figure 16 to Figure 19 present the tracking performance of the RMRCTC during simultaneous joint movements. Figure 16 shows the input (Trajec. Joint #) and output (Output Joint #) trajectories, demonstrating that the output trajectories closely followed the input trajectories.

Figure 17 illustrates the tracking errors that occurred during simultaneous joint movements, with all joints starting simultaneously and tracking the input trajectories. The maximum tracking errors recorded were [0.38°, 0.65°, 0.22°, 0.33°, 0.13°, 0.16°, 0.17°].

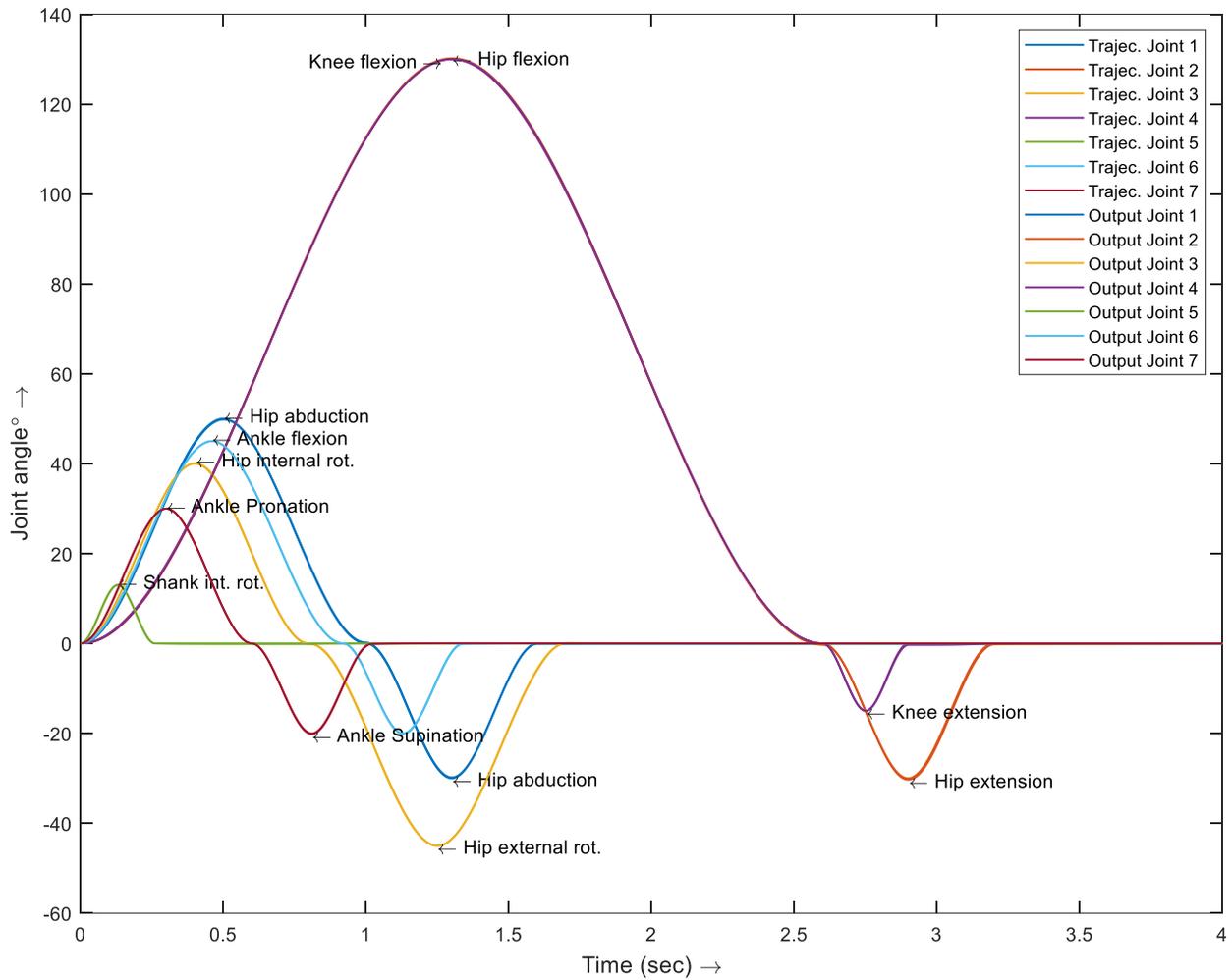

Figure 16 Simulation result for simultaneous joint movement using MRCTC

Figure 18 illustrates the joint torque required by both the model and the plant during simultaneous joint movements. Similar to sequential joint movements, the plant required more torque than the model for simultaneous movements.

The torque required by the plant consists of the torque needed by the model, along with additional torque to compensate for the relative errors between the input trajectory and the plant's output.

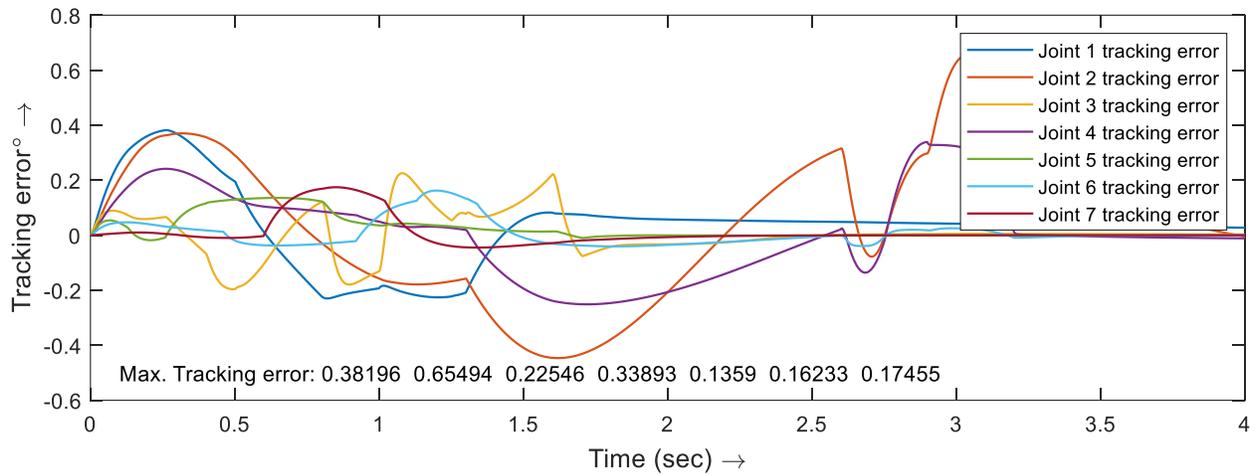

Figure 17 Tracking error during simultaneous joint movement using MRCTC

Figure 19 shows the friction torque generated during simultaneous joint movements. Similar to the sequential joint movements joint 2 experienced the highest friction torque, followed by joints 1, 4, and the others.

All the PID controller gains used to make the plant behave like the model were determined through trial and error. Fine-tuning the controller gains may further reduce the tracking error.

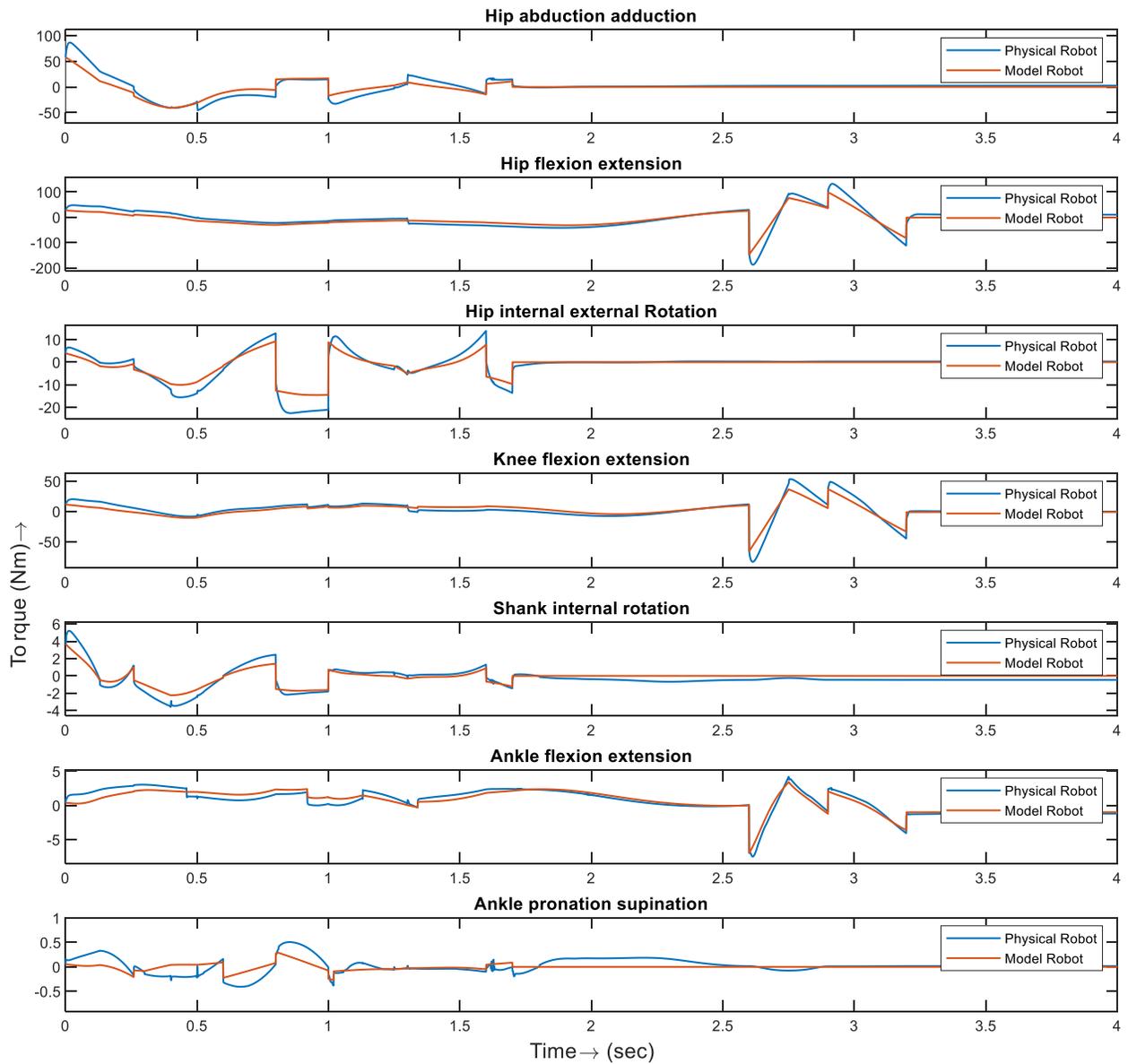

Figure 18 Joint Torque Requirements for the Model and Physical Robot during Simultaneous Joint Movement with RMRCTC

Figure 19 shows the joint friction torque developed during the simultaneous joint movements.

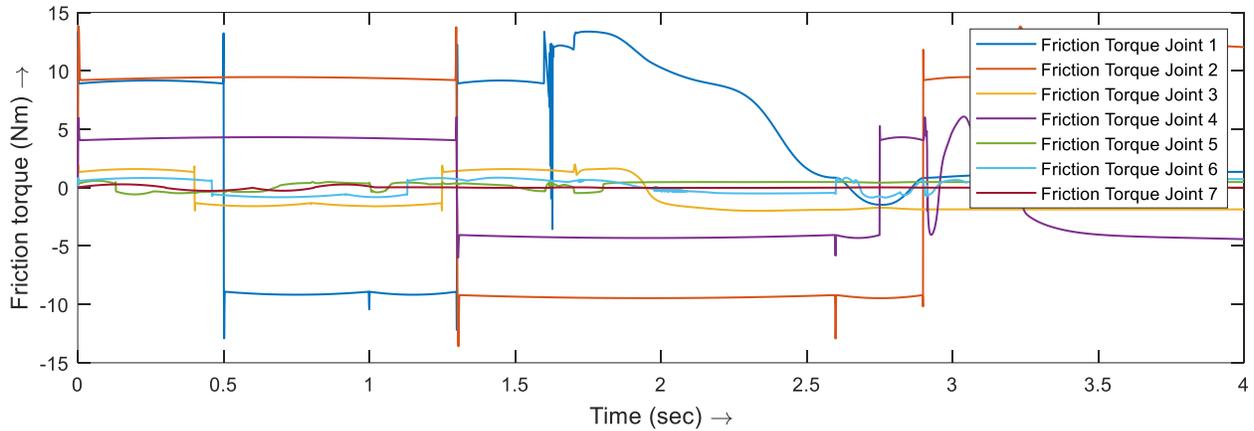

Figure 19 Friction Torque During Simultaneous Joint Movement

## 6. Controller performance verification

Statistical analysis was conducted to assess the controller's robustness and tracking performance under extreme conditions. Of the dynamic model's various parameters, only two were variable: the subject's weight and height. These two parameters were adjusted (as shown in Table 6), and different values were applied while collecting the resulting trajectory tracking errors.

Table 6: Weight and Height used for statistical analysis

| Subject's Weight | Subject's Height |
| --- | --- |
| 150 lbs. | 50 inch |
| 160 lbs. | 55 inch |
| 170 lbs. | 60 inch |
| 180 lbs. | 65 inch |
| 190 lbs. | 70 inch |
| 200 lbs. | 75 inch |

Figure 20 illustrates the joint trajectory tracking errors for various weights and heights of the subject. The error histogram reveals that changes in the subject's weight have negligible impact on the trajectory tracking errors, while variations in height do affect them. The median and standard deviation of the joint trajectory tracking errors were also calculated and maximum possible trajectory tracking error were tabulated in Table 7.

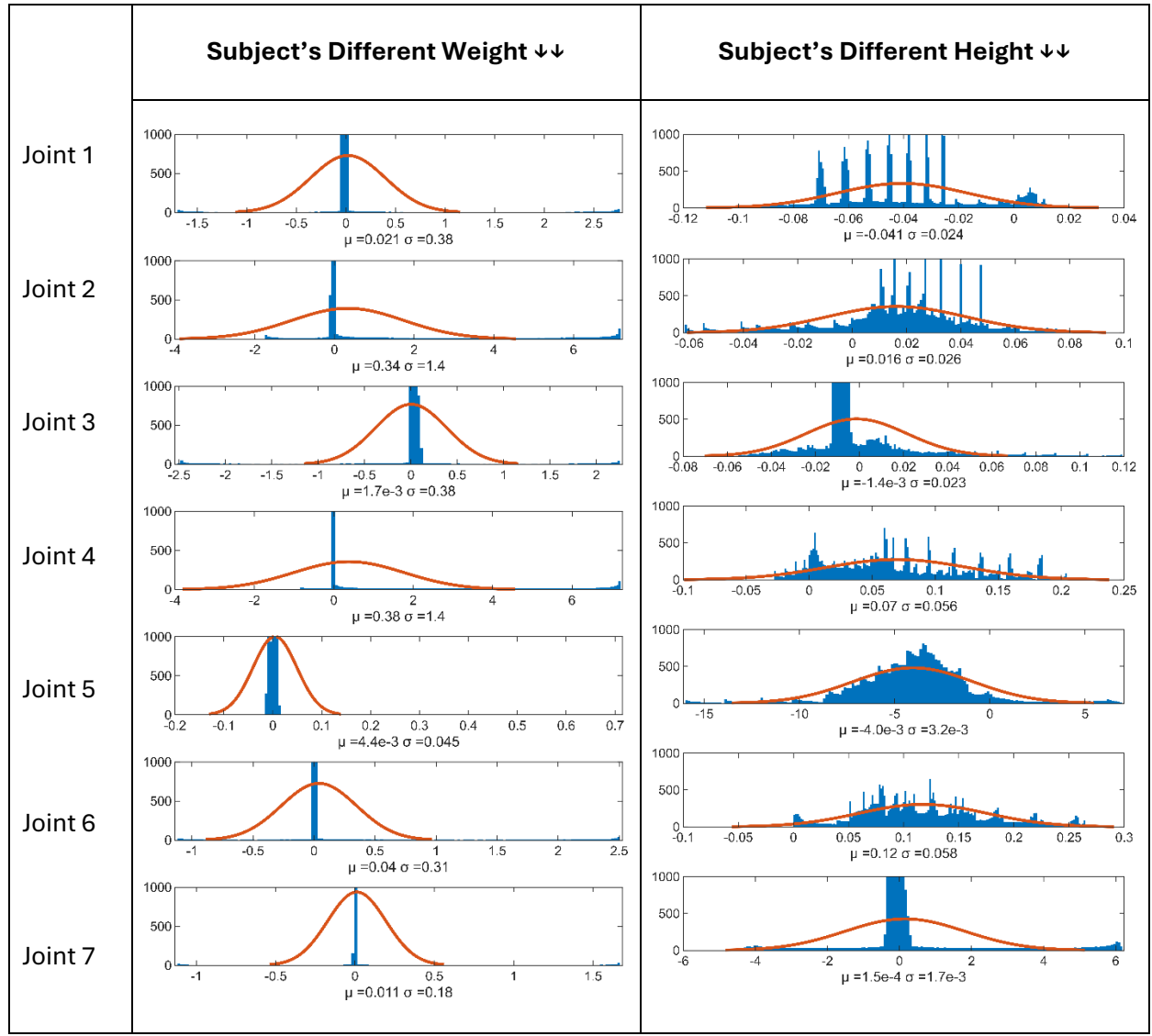

Figure 20 Joint tracking error histogram based on subject's weight and height variations

Table 7 Mean and variance of the trajectory tracking errors

| Trajectory tracking Error | Weight | | Max Error (99.70% coverage) | Height | | Max Error (99.70% coverage) |
|---|---|---|---|---|---|---|
| | Median ($\mu$) | STD ($\sigma$) | ($\mu \pm 3\sigma$) | Median ($\mu$) | STD ($\sigma$) | ($\mu \pm 3\sigma$) |
| Joint 1 | 0.021 | 0.38 | 1.16 | -0.041 | 0.024 | 0.113 |
| Joint 2 | 0.34 | 1.4 | 4.54 | 0.016 | 0.026 | 0.094 |
| Joint 3 | $1.7 \times 10^{-3}$ | 0.38 | 1.14 | $-1.4 \times 10^{-3}$ | 0.023 | 0.070 |
| Joint 4 | 0.38 | 1.4 | 4.58 | 0.07 | 0.056 | 0.238 |
| Joint 5 | $4.4 \times 10^{-3}$ | 0.045 | 0.13 | $-4 \times 10^{-3}$ | 0.023 | 0.073 |
| Joint 6 | 0.04 | 0.31 | 0.97 | 0.07 | 0.056 | 0.238 |
| Joint 7 | 0.011 | 0.18 | 0.55 | $-4.0 \times 10^{-3}$ | $3.2 \times 10^{-3}$ | 0.0136 |

Based on the obtained mean/median and the standard deviation the maximum possible trajectory tracking errors were calculated. Syms like joint 2 and joint 5 have higher standard deviation that may lead to tracking errors up to 5°. It can also be concluded that based on the subject's height and weight the controller's gain can be tuned to get better accuracy.

## Conclusions

A 7-DoF kinematic and dynamic model of the human lower extremity was developed using the Lagrange energy method. The performance of the proposed RMRCTC controller was assessed through dynamic simulation, with a realistic friction model incorporated to simulate a real robot. The controller demonstrated excellent tracking performance for both sequential and simultaneous joint movements. Trajectory tracking error bounds were observed based on the statistical analysis. Future work involves the realization of RMRCTC for exoskeleton robot control.

[40]	E. Pennestrì, V. Rossi, P. Salvini, and P. P. Valentini, "Review and comparison of dry friction force models," *Nonlinear Dynamics,* vol. 83, pp. 1785-1801, 2016/03/01 2016.

[41]	J.-L. Ha, R.-F. Fung, C.-F. Han, and J.-R. Chang, "Effects of Frictional Models on the Dynamic Response of the Impact Drive Mechanism," *Journal of Vibration and Acoustics,* vol. 128, pp. 88-96, 2005.

[42]	S. K. Hasan and A. K. Dhingra, "Development of a model reference computed torque controller for a human lower extremity exoskeleton robot," *Proceedings of the Institution of Mechanical Engineers, Part I: Journal of Systems and Control Engineering,* vol. 235, pp. 1615-1637, 2021/10/01 2021.

# List of Figures

# List of Tables